\documentclass[10pt,twocolumn,letterpaper]{article}

\usepackage{wacv}
\usepackage{times}
\usepackage{epsfig}
\usepackage{graphicx}
\usepackage{amsmath}
\usepackage{amssymb}
\usepackage{booktabs}
\usepackage{multirow}

\usepackage[accsupp]{axessibility}
\newcommand\scalemath[2]{\scalebox{#1}{\mbox{\ensuremath{\displaystyle #2}}}}

\def\wacvPaperID{242} 

\wacvfinalcopy 

\ifwacvfinal
\usepackage[breaklinks=true,bookmarks=false]{hyperref}
\else
\usepackage[pagebackref=true,breaklinks=true,colorlinks,bookmarks=false]{hyperref}
\fi

\begin{document}

\title{Understanding the Role of Mixup in Knowledge Distillation: \\An Empirical Study}

\author{Hongjun Choi, Eun Som Jeon, Ankita Shukla, Pavan Turaga\\ \\
Geometric Media Lab\\
School of Arts, Media and Engineering, Arizona State University\\
School of Electrical, Computer and Energy Engineering, Arizona State University\\
{\tt\small hchoi71@asu.edu, ejeon6@asu.edu, ashukl20@asu.edu, pturaga@asu.edu}
}
\maketitle
\thispagestyle{empty}

\begin{abstract}
Mixup is a popular data augmentation technique based on creating new samples by linear interpolation between two given data samples, to improve both the generalization and robustness of the trained model. Knowledge distillation (KD), on the other hand, is widely used for model compression and transfer learning, which involves using a larger network's implicit knowledge to guide the learning of a smaller network. At first glance, these two techniques seem very different, however, we found that ``\textit{smoothness}" is the connecting link between the two and is also a crucial attribute in understanding KD's interplay with mixup. Although many mixup variants and distillation methods have been proposed, much remains to be understood regarding the role of a mixup in knowledge distillation. In this paper, we present a detailed empirical study on various important dimensions of compatibility between mixup and knowledge distillation. We also scrutinize the behavior of the networks trained with a mixup in the light of knowledge distillation through extensive analysis, visualizations, and comprehensive experiments on image classification. Finally, based on our findings, we suggest improved strategies to guide the student network to enhance its effectiveness. Additionally, the findings of this study provide insightful suggestions to researchers and practitioners that commonly use techniques from KD. Our code is available at \url{https://github.com/hchoi71/MIX-KD}. \end{abstract}

\section{Introduction}
Deep neural networks have achieved impressive performance on a wide range of tasks including language translation \cite{vaswani2017attention, tan2019multilingual}, image classification \cite{xie2020self, he2019bag}, and speech recognition \cite{chiu2018state, pereyra2017regularizing}. To further improve the model's efficiency and performance, a large number of training techniques have been proposed such as mixup-augmentation \cite{zhang2017mixup, yun2019cutmix} and knowledge distillation \cite{hinton2015distilling}. In specific, mixup \cite{zhang2017mixup} is a commonly used data augmentation technique based on using convex combinations of samples, and their labels. This technique was introduced to improve generalization, as well as increase the robustness against adversarial examples. Recently, there has been an increasing interest in reducing the model size while preserving comparable performance, which narrows the gap between large networks and small networks. KD is one of the promising methods for this demand \cite{hinton2015distilling}. The goal of KD  is to exploit the ability to learn concise knowledge representation (logit or feature) from a larger model and then embed such knowledge into a smaller model. For example, in image classification, deep neural networks produce class probability by using softmax function that converts logit $f_{i}$ into a probability $p_{i}$ by comparing $f_{i}$ with other logits as follows: $p_{i}=\frac{\exp(f_{i}/\mathcal{T})}{\sum_{j}\exp(f_{j}/\mathcal{T})}$. In the conventional knowledge distillation \cite{hinton2015distilling}, the temperature $\mathcal{T}$ is utilized to generate a softer distribution of pseudo-probabilities among the output classes, where a higher temperature increases the entropy of the output, thus, providing more information to learn for the student model.

\begin{figure*}[t!]
	\centering
	\includegraphics[width=0.67\linewidth]{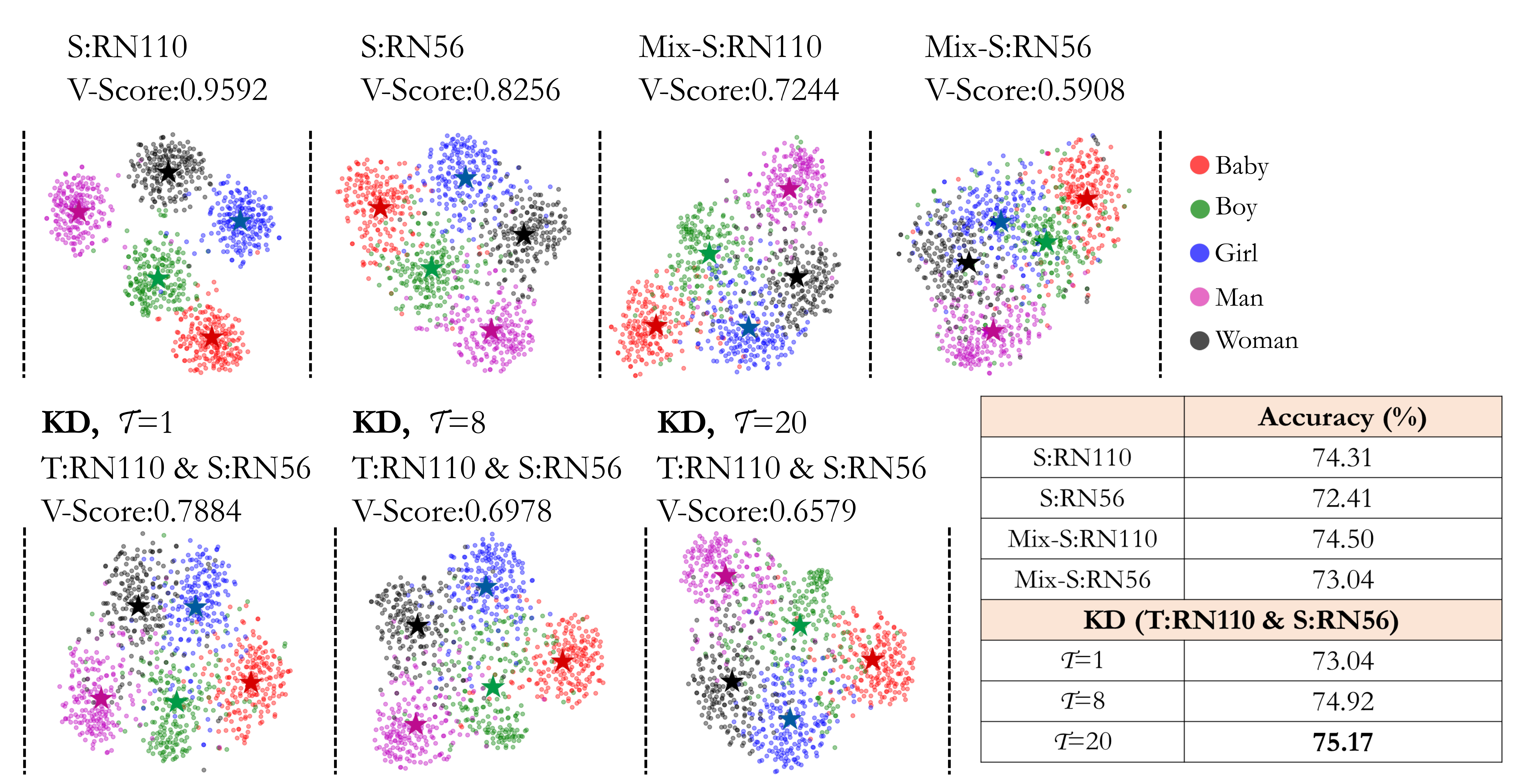}
	\caption{Feature representation of CIFAR100 extracted from the penultimate layer. Here, we only illustrate the train set as the student and teacher networks are trained on the same train set. In the first row, we observe that the higher capacity model (ResNet110 shown as RN110) promotes a tighter clustering in each class (higher V-score) than ResNet56 (RN56). Meanwhile, mixup-trained models (Mix-S:RN110 and Mix-S:RN56) disperse features of the same instances. The second row of the figure shows that increasing temperature $\mathcal{T}$ in KD has a similar effect on the projections, resulting in dispersed feature representations. Here, $\star$ indicates the mean point of each cluster. The table shows the corresponding test accuracy. Best viewed in color.}\label{Intro_smooth} 
\end{figure*}

\textbf{Motivation: } On the surface, mixup and KD are very different, however, we found that ``\textit{smoothness}" turns out to be the connecting link between the two and a very important attribute to understanding KD's interplay with mixup. Intuitively, the KD involves a student mimicking smoothed probability distribution of the teacher, whereas mixup artificially introduces smoothness (the labels are not strictly one-hot) in a model via linear combinations of the labels. Although mixup-augmentation and KD are common techniques in training networks in various applications, the interplay between the two has not been well explored. In this paper, we investigate in detail the impact of mixup-augmentation on knowledge distillation.

To develop insight into several interesting behaviors of the network, we provide various visualizations in feature and logit levels. For example, to analyze how mixup enforces a feature representation in-between each class, we pick 5 semantically similar classes from the CIFAR100 dataset (baby, boy, girl, man, and woman). Then, we project the features extracted from the penultimate layer into 2-D using tsne \cite{van2008visualizing} as seen in Figure \ref{Intro_smooth}. The first row of the figure represents the feature representation for two networks trained from scratch with mixup (Mix-S:ResNet110 or Mix-S:ResNet56) and without mixup (S:ResNet110 or S:ResNet56). Generally, we observe that the higher capacity model (S:ResNet110) encourages the deep networks to learn tight projections in each class while the lower capacity model (S:ResNet56) learns more diffuse projections. This is also verified by clustering metrics, such as the V-score \cite{rosenberg2007v}, where a higher value implies better clustering. Interestingly, the mixup-trained models disperse features in similar classes, compared to the one trained without mixup, even though a mixup-trained model shows better test accuracy due to an increase in generalization on unseen data. Meanwhile, we found similar observations on the feature representation from distilled models when a higher temperature is used (i.e., distilling more softer logit) as seen in the second row of Figure \ref{Intro_smooth}. From the table in Figure \ref{Intro_smooth}, high temperature $\mathcal{T}$ is a default choice to improve the performance. In this way, the teacher transfers more information to the student even though higher temperatures superficially promote feature dispersion in similar classes. However, transferring a high quality of supervision to a student network is also crucial as it can guide the student to learn discriminative representations from the superior teacher network. Thus, we seek a way to increase supervisory signals from teacher to student, without compromising on performance. This is where mixup presents a way forward. 


In short, the crux of our insight is that if a teacher is trained with smoothed data from mixup, then further `smoothing' at high temperatures during distillation can be avoided. This ensures stronger supervisory signals while enjoying the benefits of dataset augmentation via mixup. We summarize our contributions as follows:

\begin{enumerate}
   \item We provide new insights into devising improved strategies for learning a student model, by a deeper understanding of the behavior of features/logits and networks trained with mixup during KD.
   \item To reduce the criticality of choosing a proper `temperature', we develop a simple rescaling scheme that brings different statistical characteristics in logits between teacher and student to a similar range while preserving the relative information between classes, thereby the temperature $\mathcal{T}$ is no longer used.
   \item We identified that strongly interpolated mixup pairs impose extra smoothness on the logits, thus we can generate only a few mixup pairs in a batch, called partial mixup (PMU), and yet achieve comparable and even better performance with this variant in KD. 
\end{enumerate}

\section{Background}\label{Background} We first introduce the background of mixup and KD through a simple mathematical description. Given the training data $\mathcal{D}=\{(x_{1},y_{1}),...,(x_{n},y_{n})\}$, the goal of the classification task is to learn classifier $f:\mathcal{X}\rightarrow\mathbb{R}^{k}$ by mapping input $x \in \mathcal{X} \subseteq \mathbb{R}^{d}$ to label $y \in \mathcal{Y}=\{1,2,...,K\}$. Let $\mathcal{L}(f(x), y)$ be the loss function that measures how poorly the classifier $f(x)$ predicts the label $y$. 

\noindent\textbf{Mixup Augmentation \cite{zhang2017mixup}} In mixup augmentation, two samples are mixed together by linear interpolation as follows: $\tilde{x}_{ij}(\lambda)=\lambda x_{i} + (1-\lambda)x_{j}$, and $\tilde{y}_{ij}(\lambda)=\lambda y_{i} + (1-\lambda)y_{j}$, where $\lambda\in [0,1]$ follows the distribution $P_{\lambda}$ where $\lambda \sim \text{Beta($\alpha$, $\alpha$)}$. Then, the mixup loss function can be described as  
\begin{equation}\label{eq:mixup_eq}
\scalemath{0.99}{\mathcal{L}_{mix}(f)= \frac{1}{n^{2}}\sum_{i=1}^{n}\sum_{j=1}^{n}\mathbb{E}_{\lambda\sim P_{\lambda}}[\mathcal{L}(f(\tilde{x}_{ij}(\lambda)), \tilde{y}_{ij}(\lambda))]},
\end{equation}
\noindent where $\mathcal{L}$ represents the cross-entropy loss function in this study. Specifically, a hyper-parameter $\lambda$ in Equation \ref{eq:mixup_eq} is used to specify the extent of mixing. In other words, the control parameter $\alpha$ in a beta distribution commands the strength of interpolation between feature-target pairs, i.e., the high $\alpha$ generating strongly interpolated samples. 

A fair amount of variants of mixup have been proposed \cite{verma2019manifold, yun2019cutmix, kim2021co}. The general strategy of these mixing-based methods is intrinsically similar in that they extend the training distribution by blending the images and mixing their labels proportionally. Thus, in this study, we only pay attention to the conventional mixup to investigate the interplay between mixup and knowledge distillation \cite{zhang2017mixup}.  

\noindent\textbf{Knowledge Distillation \cite{hinton2015distilling}} In knowledge distillation, given a pre-trained teacher model $f^{T}$ on the dataset in advance, the student model $f^{S}$ is trained over the same set of data by utilizing concise knowledge generated by $f^{T}$. In specific, once the teacher network is trained, its parameter is frozen during the training in KD, and then, the student network is trained by minimizing the similarity between its output and the soft labels generated by the teacher network. To this end, we minimize the divergence between logits of the student and the teacher as follows:
\begin{equation}\label{eq:kd_eq}
\scalemath{0.99}{\mathcal{L}_{kd}(f^{T},f^{S})=\frac{1}{n}\sum_{i=1}^{n}KL(\mathtt{S}(\frac{f^{T}(x_{i})}{\mathcal{T}}), \mathtt{S}(\frac{f^{S}(x_{i})}{\mathcal{T}}))}
\end{equation}
\noindent where $\mathtt{S}$ indicates the softmax function, $KL$ measures the Kullback-Leiber divergence, and output logits of the model smoothen by temperature, $\mathcal{T}$.

\noindent\textbf{Data Augmentation in Knowledge Distillation} Recently, several works have utilized data augmentation and achieved promising results in that augmented samples enable the networks to learn relaxed knowledge from different views in distillation frameworks \cite{wang2020knowledge, wang2020neural, li2021smile}. Across the broadly available methods, they often do not provide insight into the inner working of the models. Meanwhile, recent works \cite{muller2019does, shen2021label} have studied KD's compatible/incompatible views with label smoothing \cite{szegedy2016rethinking} to provide such insight into deep models via empirical analysis. Instead, we are interested in the underlying mechanism of augmentation in the light of the KD process. Our study spans double aspects that attempt to present both compatible and incompatible views through comprehensive empirical analysis. Further, based on the observations, we suggest a better learning strategy to enhance the network's performance.

\begin{figure}[t!]
\vspace{-1.0em}
\includegraphics[width=7.5cm]{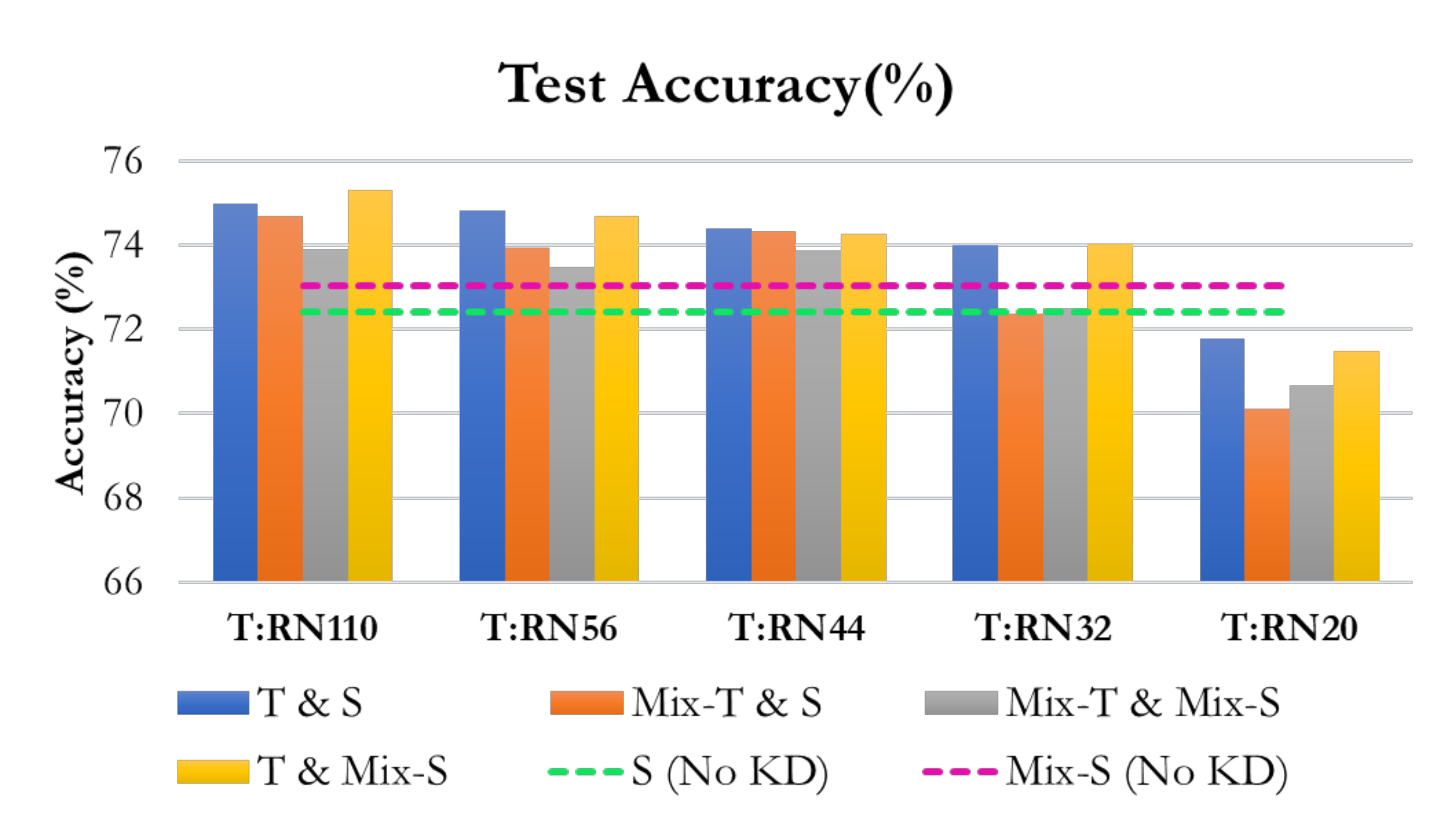}
\caption{The student (fixed RN56) test accuracy on CIFAR100 under different teacher networks with $\mathcal{T}=4$. Here, T and S denote the standard teacher and the standard student. Mix-T and  Mix-S are mixup-trained teachers and mixup-trained students, respectively. Mixup-trained teachers guide lower-performance students than the one with standard teacher models, and lower-capacity teacher models generally distill inferior students.}\label{Motiv_Acc}
\vspace{-1.0em}
\end{figure} 

\section{Key findings from Mixup and KD interplay}\label{closer_look}
In this section, we discuss our main findings. We first begin by referring to Figure \ref{Motiv_Acc}. Here, we describe four possible scenarios where mixup-augmentation could be involved in KD as follows; standard teacher and standard student (T\&S), mixup-trained teacher and standard student (Mix-T\&S), mixup-trained teacher and mixup-trained student (Mix-T\&Mix-S), and standard teacher and mixup-trained student (T\&Mix-S) under the same temperature being $\mathcal{T}=4$. We fix the student model as ResNet56 (RN56) and evaluate it by varying the teacher models from ResNet20 (RN20) to ResNet110 (RN110).

\begin{figure*}[h!]
	\centering
	\includegraphics[width=0.93\linewidth]{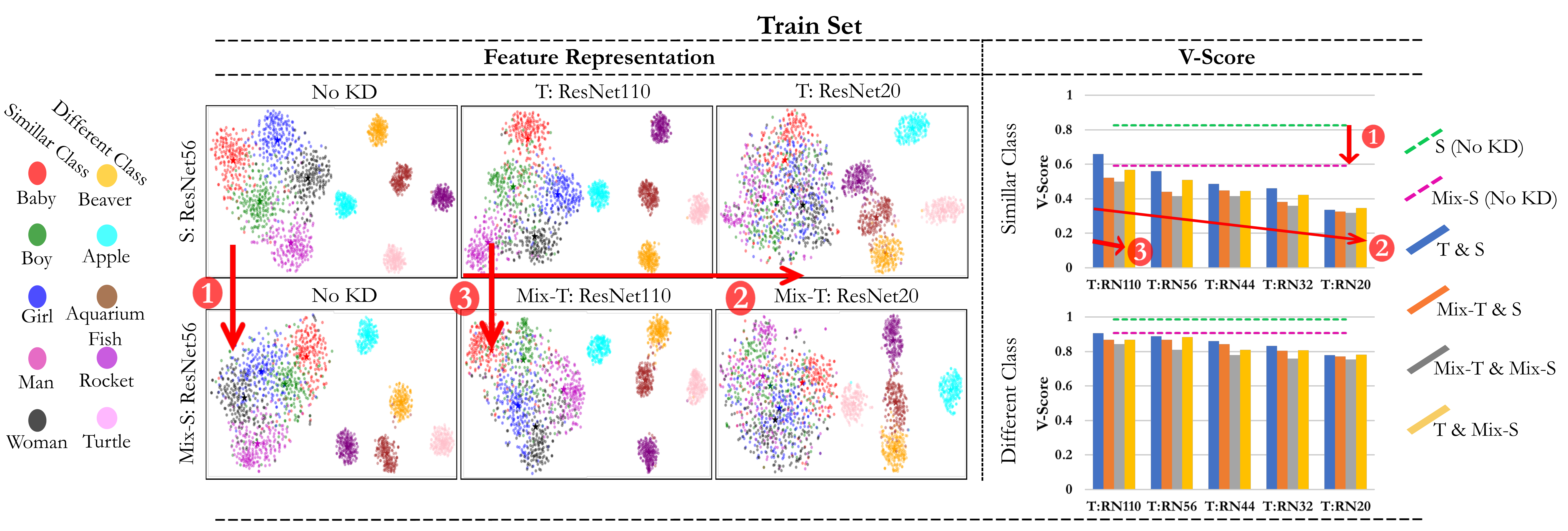}
	\caption{Feature representations of the penultimate layer with various combinations (T:standard teacher, S:standard student, Mix-T:mixup-trained teacher, Mix-S:mixup-trained teacher) and the corresponding V-scores on the CIFAR100 dataset. We selected two groups, semantically similar classes (Baby, Boy, Girl, Man, and Woman), and semantically different classes (Beaver, Apple, Aquarium Fish, Rocket, and Turtle). Here, the features are extracted from the student networks. \textbf{Observation 1) } Mixup augmentation encourages the features of samples from similar classes to be dispersed, while it still well-preserves the feature separability in different classes. \textbf{Observation 2) } The use of a lower-accuracy teacher network enables the student to learn by less-discriminative features, degrading the performance of KD. \textbf{Observation 3) } In presence of a mixup-trained teacher, the dispersion of feature representation undermines the benefit of supervision by the teacher. $\star$ represents the center point of each cluster. Best viewed in color.}\label{Motiv_Qual_Quant} 
\vspace{-1.0em}
\end{figure*}

As seen in Figure \ref{Motiv_Acc}, we can make two observations: First, the student performance with the help of a mixup-trained teacher (Mix-T\&S, Mix-T\&Mix-S) always shows less effectiveness than with the help of a standard teacher (T\&S, T\&Mix-S) under the same settings of temperature even if the mixup-trained teacher itself shows better test accuracy than the teacher trained without mixup. Second, generally, higher capacity teacher models distill better students, but with a lower capacity teacher (T:RN44 \& S:RN56), the student performance still shows improvement in accuracy, compared to the vanilla student model (green dashed line, S (No KD)). Based on these observations, this paper aims to investigate the following questions: 1) \textit{Why does a mixup-trained teacher model impair the student's effectiveness in KD?} We answer this question in section \ref{closer_look}. Then, 2) \textit{what can we do to improve the effectiveness of knowledge distillation when mixup-augmentation is applied?} We address this question in section \ref{std_partialmixup}.

\noindent\textbf{Observation 1) Mixup vs Non-Mixup.} To investigate the effect of the network trained with mixup-augmentation, we selected a few classes and divided them into two groups: (1) semantically different classes (Beaver, Apple, Aquarium Fish, Rocket, and Turtle) and (2) semantically similar classes (Baby, Boy, Girl, Man, and Woman), all from CIFAR100. For example, in Figure \ref{Motiv_Qual_Quant}, the left figure illustrates the feature representation of the penultimate layer on train sets. If we look at the number 1 in the red circle, the projections of a mixup-trained model from similar classes are more dispersed while the ones of different classes are still well-preserved in their structure. This information loss can also be measured by clustering metric, V-score on the right histogram of the figure, resulting in a drastic drop in V-score for similar classes.

\noindent\textbf{Observation 2) Distilling from low-accuracy teacher models.} If the student network is settled, retaining the fine quality of supervision (i.e., the performance of a teacher network) is crucial in training better students. As followed in the number 2 in the red circle, the projections of the student with the lower-accuracy teacher are notably dispersed, significant drop in V-score, eventually impairing the student's performance. This implies that the better student is distilled with the help of discernible features given by the high-capacity teacher. 

\noindent\textbf{Observation 3) Distilling from a mixup-trained teacher network.} Observation 1) showed that the mixup-trained model scatters the features in similar classes. We now look at the case where a mixup-trained teacher conveys the knowledge to the student. In KD, as the student and teacher models are trained on the same train set, we argue that a student trained with supervision by a mixup-trained teacher cannot take advantage of learning superior knowledge due to feature scattering. As shown by number 3 in this figure, feature representations of similar classes in the student network eventually become more spread out, and V-scores also drop on both train and test sets. We provide the visualizations of train and test sets in the supplementary material. 

\begin{figure*}[h!]
	\centering
	\includegraphics[width=0.82\linewidth]{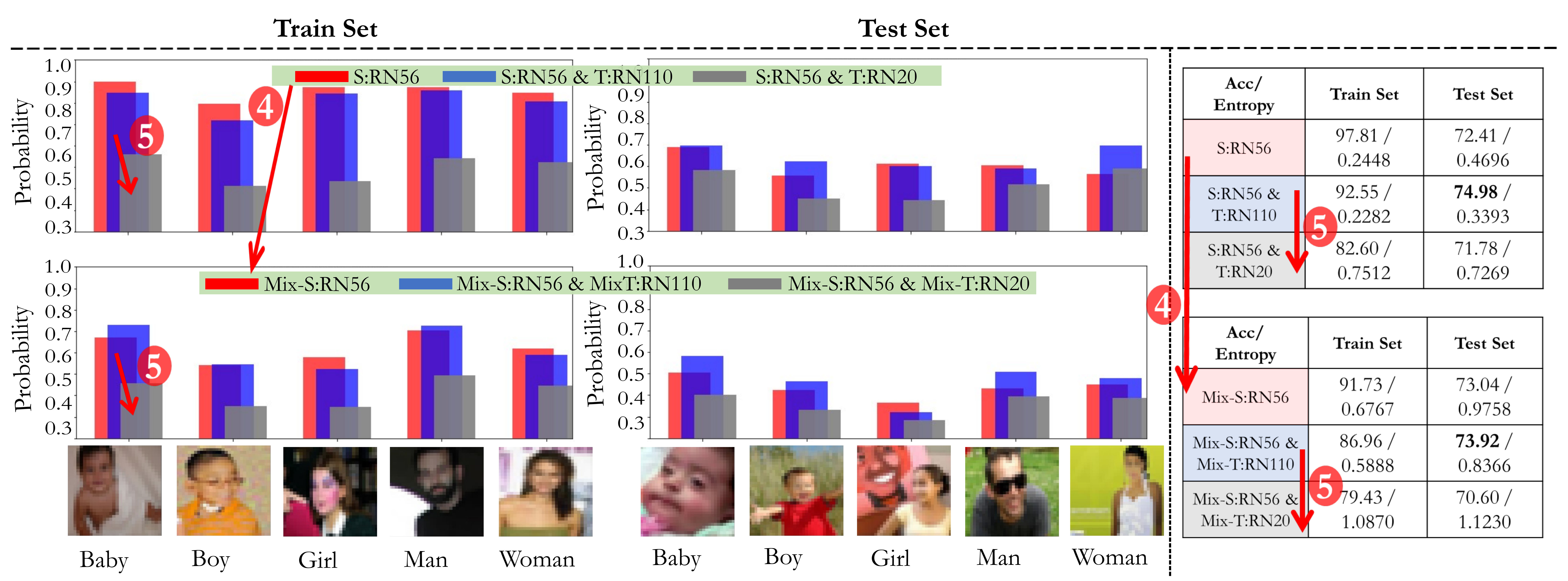}
	\caption{Probability distributions of the following configurations: with/without a mixup-trained model (S:RN56/Mix-S:RN56), the standard student model trained with the help of different capacity teachers (T:RN110 \& S:RN56, T:RN20 \& S:RN56), and the student model trained with mixup in the presence of different mixup-trained teachers (Mix-T:RN110 \& Mix-S:RN56, Mix-T:RN20 \& Mix-S:RN56). We show the mean distributions for five similar classes in CIFAR100 on the left and provide the average accuracy and entropy values for all examples. \textbf{Observation 4)} When the model is trained with the mixup, the confidence in predictions decreases on both sets (also verified by higher entropy value). \textbf{Observation 5)} Also, in the case of distilling from lower-capacity teachers, the confidence in predictions is much lower than the distilled model from high-capacity teachers. Best viewed in color.}\label{Motiv_Logit} 
\vspace{-1.0em}
\end{figure*}

\noindent\textbf{Observation 4) and 5) Logit representation.} Unlike the feature representations shown in observations 1)-3), we further visualize the probability distribution of the student network. First, we average probabilities of all classes on the train and test of CIFAR100 and illustrate the mean distribution of the examples that belong to the same category to show the model's prediction of that category. To compare with quantitative measurement, we also provide the average accuracy and entropy values computed across all examples in Figure \ref{Motiv_Logit}, where the entropy is popularly used to measure the smoothness of the distribution. The higher entropy value is the smoother distribution.

Here, we observe two intriguing phenomena; Observation 4) A mixup-trained model produces softer output logits, illustrated by the short red bars in both train and test set. From this observation, we surmise that a mixup-augmentation involved in training the student in KD contributes extra smoothness to the logits. Further, the student learned from a standard teacher outperforms the mixup-trained teacher in terms of accuracy (74.98\% vs 73.92\% and 71.78\% vs 70.60\%) under the same settings of the temperature $\mathcal{T}=4$. Conclusively, when distilling from a mixup-trained teacher, the use of high temperature adversely impacts the accuracy of the student. Observation 5) When the inferior knowledge produced by the low-accuracy teacher (T:RN20) transfers to the student, the confidence level on the prediction significantly falls as seen in the gray bars on both sets, leading to a remarkable decrease in test accuracy on both cases (74.98\% $\rightarrow$ 71.78\%, 73.92\% $\rightarrow$ 70.60\%). It implies that transferring decent quality of supervision to the student is crucial to successful knowledge distillation.

\noindent\textbf{Observation 6) Mix-T\&S at low temperature.} From observation 5), while increased $\mathcal{T}$ is believed to be helpful to produce better representation for KD, we remark that in the presence of a mixup-trained teacher, an increased $\mathcal{T}$ can be an adverse effect on the performance of KD because of the feature dispersion/exorbitant smoothness in the logit. At this point, one might give rise to the following question. \textit{What if we lower the temperature to lessen smoothness in the logit?}

\begin{figure}[h!]
	\centering
	\includegraphics[width=1.0\linewidth]{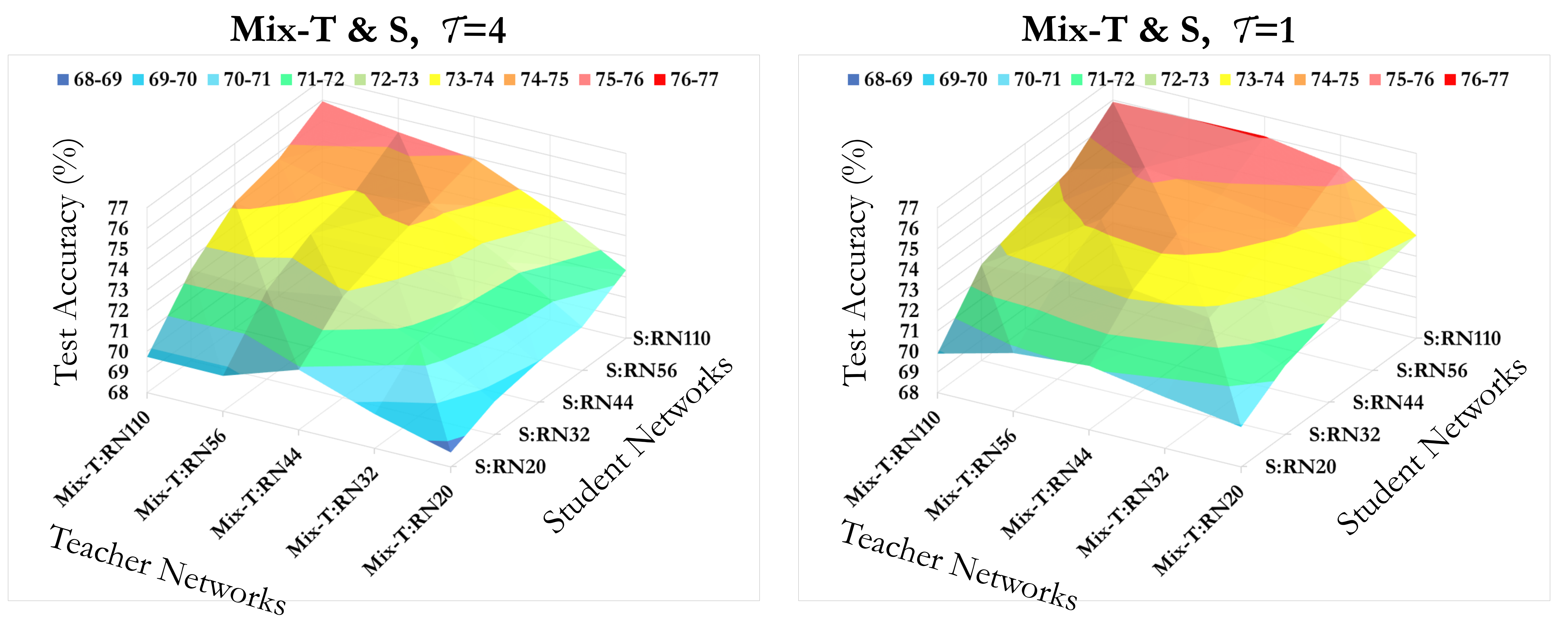}
	\caption{Visualization with the heatmap representation. When distilling from a mixup-trained teacher (Mix-T) at a higher temperature, it transfers inferior knowledge to the student. With lower temperature $\mathcal{T}=1$, it produces better students, covering the relative high-accuracy region in the heatmap.}\label{test_acc_heatmap} 
\vspace{-1.0em}
\end{figure}

To investigate how much temperature impacts the test performance in KD, we visualize the test accuracy as a heatmap in Figure \ref{test_acc_heatmap}. In this figure, we explore all combinations of a mixup-trained teacher and student. The left panel of this figure shows test accuracy under the setting $\mathcal{T}=4$ and the right panel is for $\mathcal{T}=1$. We note that when a lower temperature, that makes the logit less smooth, is applied, the better the quality of knowledge that is transferred to the student. This result demonstrates that knowledge from a mixup-trained teacher at high temperatures renders the student model less effective.

\section{Effective learning strategies for Mix-KD}\label{std_partialmixup} Based on our findings, we observe that smoothness in the logit is a critical attribute in distillation, as seen in Figure \ref{test_acc_heatmap}. Since it is difficult to measure suitable smoothness from the perspective of the student, many distillation methods have relied heavily on a naive brute-force search to find the proper temperature. To reduce the criticality of choosing the proper temperature and alleviating excessive smoothness by a strong mixup, we introduce an effective learning strategy to improve the performance of KD.

\noindent\textbf{Partial mixup.} The control parameter in the beta distribution plays a key role in controlling the strength of interpretation, which also affects the degree of softened output. As $\alpha \rightarrow 1$, it provides softer output logits. However, there is a trade-off between avoiding excessive smoothness and improvement in robustness against adversarial attacks in knowledge distillation. To alleviate this issue, we suggest generating only small amounts of mixup pairs used in training, called partial mixup (PMU). For example, PMU=10\% refers to only 10\% mixup pairs used in a batch and the rests keep untouched. To further understand the behavior of how partial mixup affects output probability, we provide toy examples using 2 classes in the supplementary material.

\noindent\textbf{Rescaled logits.} Here, we suggest using standard deviation as temperature such that the logit produced by the output layer of the network is rescaled by dividing it by the standard deviation of that logit, thereby the temperature $\mathcal{T}$ is no longer hyper-parameter. This way can bring two different statistical characteristics between the teacher and the student logits to similar ranges while it does not hurt the relative structure in-between classes. We underline the importance of rescaling logit since the random mixing portion $\lambda$ from a beta distribution governs the smoothness degree, yielding irregular smooth output at each iteration. Therefore, we replace the output logit with rescaled one, and the loss function becomes as follows:
\begin{equation}\label{eq:std_eq}
\scalemath{0.87}{\mathcal{L}_{kd\_r}(\tilde{f}^T,\tilde{f}^S)=\frac{1}{n}\sum_{i=1}^{n}KL(\mathtt{S}(\frac{f^{S}(x_{i})}{\sigma(f^{S}(x_{i}))}), \mathtt{S}(\frac{f^{T}(x_{i})}{\sigma(f^{T}(x_{i}))}))},
\end{equation}
\noindent where $\tilde{f} =f/\sigma(f)$, $\mathtt{S}$ indicates the softmax function, $KL$ is the Kullback-Leibler divergence, and $\sigma(\cdot)$ is standard deviation of input logit. Then, the final training objective for the student in KD is as follows:
\begin{equation}\label{eq:final_loss_eq}
\scalemath{0.91}{\min\mathbb{E}_{(x,y)\sim\mathcal{D}}\mathbb{E}_{\lambda\sim P_{\lambda}}[\gamma_{kd}\mathcal{L}_{mix}(\tilde{f}^S) + \alpha_{kd}\mathcal{L}_{kd\_r}(\tilde{f}^T, \tilde{f}^S)]},
\end{equation}
\noindent where $\gamma_{kd}$ and $\alpha_{kd}$ are balancing parameters. Note, the hyper-parameter of the partial amount in PMU is not denoted in this equation.

\begin{figure}[h!]
\vspace{-1.0em}
\centering
\includegraphics[width=0.80\linewidth]{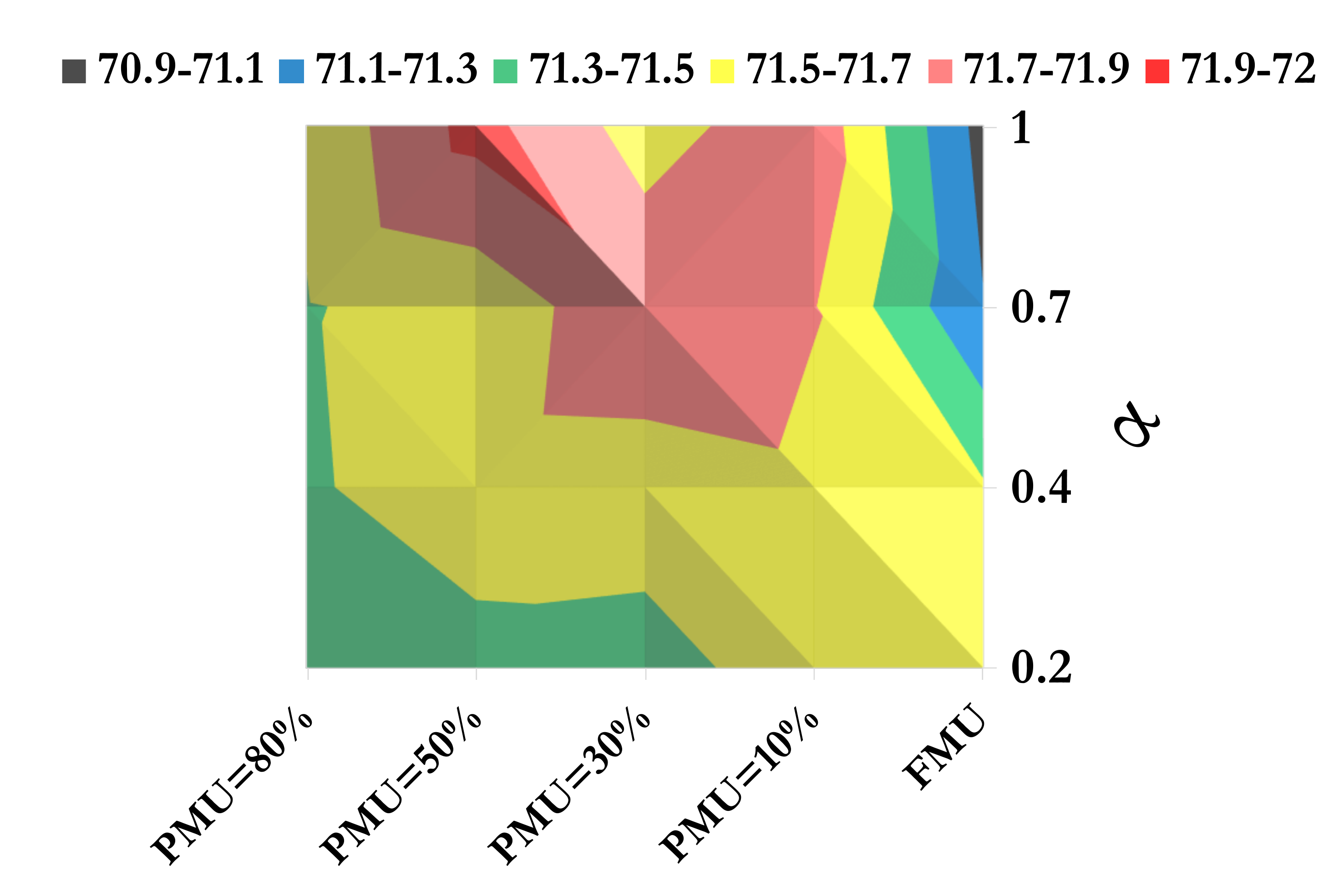}
\caption{Test accuracy on CIFAR100 under different configurations of $\alpha$ and the amounts of mixup pairs. The reported accuracy is averaged over 3 runs.}\label{choosing_param}
\vspace{-0.0em}
\end{figure}

\noindent\textbf{Choice of $\boldsymbol\alpha$ and PMU.}\label{choosing_parameter} Further, we investigate how each hyper-parameter (partial amount in PMU and the control parameter $\alpha$) affects the student performance in KD under the following combination, T:RN110 and Mix-S:RN20 on the CIFAR100 dataset. We evaluate the student performance by varying the degree of partial amount from 10\% to 80\% with different control parameters in the range of $[0.2, 0.4, 0.7, 1.0]$. Figure \ref{choosing_param} indicates that PMU with high $\alpha$ generally leads to better students in this setting. The lowest performance is observed in FMU (fully mixup pairs used) with $\alpha=1.0$. We note that PMU does not necessarily aim to outperform FMU, but it plays a regulatory role to adjust smoothness finely, thus helping us to understand the effect of smoothness in KD. We present performance analysis in section \ref{sub_image_classification} and also further study how partial mixup responds to adversarial attacks in section \ref{Ablation_study_section}.

\section{Experiments}\label{experiments}
In this section, we present the experimental results to validate our conjectures and findings. In the previous section, we noted that (1) a mixup-trained teacher generates an inferior student compared to the teacher trained without mixup under the same temperature (2) but lowering temperature can recover the test accuracy. In KD, mixup augmentation is only useful for the teacher if a mixup-trained teacher provides the student with further benefits such as better knowledge and additional power of robustness. Through our analysis, we see that the expected effect of the use of mixup-trained teachers is not satisfactory. Therefore, we use the teacher trained without mixup for our experiment, i.e., T\&Mix-S.

\subsection{Image classification on CIFAR100 \& ImageNet}\label{sub_image_classification}
\begin{table*}[h!]
\vskip -0.0in
\caption{CIFAR-100 test accuracy (\%) of student networks trained with several distillation methods. The results of the other distillation methods except $*$ are quoted from \cite{tian2019contrastive}. The best and second-best results are highlighted in \textbf{Bold} and \textcolor{red}{Red}. $*$ is performed with our implementation based on the author-provided code.}
\label{test_accuracy_table_on_cifar100}
\vspace{-0.1in}
\begin{center}
\scalebox{0.75}{\begin{tabular}{l|ccccccc|cccccr}
\toprule
Teacher & W40-2 & W40-2 & RN56 & RN110 & RN110 & RN32$\times$4 & VG13 & VG13 & RN50 & RN50 & RN32$\times$4 & RN32$\times$4 & W40-2\\
Student & W16-2 & W40-1 & RN20 & RN20 & RN32 & RN8$\times$4 & VG8 & MN2 & MN2 & VG8 & SN1 & SN2 & SN1\\
\midrule
Teacher & 75.61 & 75.61 & 72.34 & 74.31 & 74.31 & 79.42 & 74.64 & 74.64 & 79.34 & 79.34 & 79.42 & 79.42 & 75.61\\
Student & 73.26 & 71.98 & 69.06 & 69.06 & 71.14 & 72.50 & 70.36 & 64.6 & 64.6 & 70.36 & 70.50 & 71.82 & 70.50\\
\midrule
KD\cite{hinton2015distilling}    & 74.92 & 73.54 & 70.66 & 70.67 & 73.08 & 73.33 & 72.98 & 67.37 & 67.35 & 73.81 & 74.07 & 74.45 & 74.83 \\
FitNet\cite{romero2014fitnets} & 73.58 & 72.24 & 69.21 & 68.99 & 71.06 & 73.50 & 71.02 & 64.14 & 63.16 & 70.69 & 73.59 & 73.54 & 73.73 \\
AT\cite{zagoruyko2016paying}    & 74.08 & 72.77 & 70.55 & 70.22 & 72.31 & 73.44 & 71.43 & 59.40 & 58.58 & 71.84 & 71.73 & 72.73 & 73.32 \\
SP\cite{tung2019similarity}    & 73.83 & 72.43 & 69.67 & 70.04 & 72.69 & 72.94 & 72.68 & 66.30 & 68.08 & 73.34 & 73.48 & 74.56 & 74.52 \\
CC\cite{peng2019correlation}     & 73.56 & 72.21 & 69.63 & 69.48 & 71.48 & 72.97 & 70.71 & 64.86 & 65.43 & 70.25 & 71.14 & 71.29 & 71.38 \\
VID\cite{ahn2019variational}      & 74.11 & 73.30 & 70.38 & 70.16 & 72.61 & 73.09 & 71.23 & 65.56 & 67.57 & 70.30 & 73.38 & 73.40 & 73.61\\
RKD\cite{park2019relational}      & 73.35 & 72.22 & 69.61 & 69.25 & 71.82 & 71.90 & 71.48 & 64.52 & 64.43 & 71.50 & 72.28 & 73.21 & 72.21 \\
PKT\cite{passalis2018learning}   & 74.54 & 73.45 & 70.34 & 70.25 & 72.61 & 73.64 & 72.88 & 67.13 & 66.52 & 73.01 & 74.10 & 74.69 & 73.89\\
AB\cite{heo2019knowledge}   & 72.50 & 72.38 & 69.47 & 69.53 & 70.98 & 73.17 & 70.94 & 66.06 & 67.20 & 70.65 & 73.55 & 74.31 & 73.34 \\
FT\cite{kim2018paraphrasing}   & 73.25 & 71.59 & 69.84 & 70.22 & 72.37 & 72.86 & 70.58 & 61.78 & 60.99 & 70.29 & 71.75 & 72.50 & 72.03 \\
NST\cite{huang2017like}   & 73.68 & 72.24 & 69.60 & 69.53 & 71.96 & 73.30 & 71.53 & 58.16 & 64.96 & 71.28 & 74.12 & 74.68 & 74.89\\
CRD\cite{tian2019contrastive}   & 75.48 & 74.14 & 71.16 & 71.46 & 73.48 & 75.51 & \textcolor{red}{73.94} & \textbf{69.73} & 69.11 & 74.30 & 75.11 & 75.65 & 76.05\\
ICKD$^{*}$\cite{liu2021exploring}   & 75.64 & 74.18 & 71.56 & 71.29 & 73.49 & 74.78 & 73.36 & 68.61 & 68.65 & 73.43 & 74.96 & 75.34 & 76.18 \\
\midrule
Ours (No mixup)   & 75.38 & 73.70 & \textcolor{red}{71.85} & 71.61 & 73.60 & 75.46 & 72.92 & 67.37 & 67.72 & 73.10 & 73.38 & 75.06 & 75.09 \\
Ours (PMU=10\%)   & \textbf{76.06} & \textcolor{red}{74.42} & \textbf{72.09} & \textbf{71.94} & \textbf{74.07} & 76.87 & 73.60 & 68.52 & 69.55 & 74.29 & 75.89 & 77.06 & 76.78 \\
Ours (PMU=50\%)   & \textcolor{red}{75.87} & \textbf{74.69} & 71.80 & \textcolor{red}{71.78} & \textcolor{red}{73.97} & \textcolor{red}{77.13} & \textbf{74.00} & \textcolor{red}{69.14} & \textcolor{red}{69.69} & \textbf{74.61} & \textcolor{red}{76.83} & \textcolor{red}{77.60} & \textbf{77.18} \\
Ours (FMU) & 75.69 & 73.34 & 70.98 & 70.99 & 73.48 & \textbf{77.25} & 73.84 & 68.81 & \textbf{69.80} & \textcolor{red}{74.50} & \textbf{77.17} & \textbf{77.92} & \textcolor{red}{77.00} \\
\bottomrule
\end{tabular}}
\end{center}
\vskip -0.1in
\end{table*}

\begin{table*}[h!]
\vskip -0.00in
\caption{Top-1 and Top-5 accuracy (\%) on ImageNet validation dataset compared with various knowledge distillation methods.}
\label{test_accuracy_table_on_imagenet}
\vskip 0.05in
\begin{center}
\scalebox{0.85}{\begin{tabular}{l|cc|cccccc|cc} 
\toprule
& Teacher & Student & KD\cite{hinton2015distilling} & AT\cite{zagoruyko2016paying} & RKD\cite{park2019relational} & SP\cite{tung2019similarity} & CC\cite{peng2019correlation} & CRD\cite{tian2019contrastive} & Ours(PMU=10\%) & Ours(FMU) \\
\midrule
Top-1 & 73.31 & 69.75 & 70.66 & 70.70 & 70.59 & 70.79 & 69.96 & 71.17 & \textcolor{red}{71.38} & \textbf{71.82} \\
Top-5 & 91.42 & 89.07 & 89.88 & 90.00 & 89.68 & 89.80 & 89.17 & 90.13 & \textcolor{red}{90.40} & \textbf{90.63} \\
\bottomrule
\end{tabular}}
\end{center}
\vskip -0.2in
\end{table*}

\noindent \textbf{Experiments on CIFAR-100:} Table \ref{test_accuracy_table_on_cifar100} compares top-1 accuracy of various distillation methods and evaluates various network choices for teacher-student. The first two rows of the Table \ref{test_accuracy_table_on_cifar100} represent many different teacher-student combinations by utilizing the networks as follows: Wide residual networks (Wd-w) \cite{zagoruyko2016wide} where d and w represent depth and width in the networks respectively, MobileNetV2 (MN2) \cite{sandler2018mobilenetv2}, ShuffleNetV1 (SN1) \cite{zhang2018shufflenet}/ShuffleNetV2 (SN2)\cite{ma2018shufflenet}, VGG (VG) \cite{simonyan2014very}, and ResNet (RN) \cite{he2016deep}. 

All the models are trained for 240 epochs with a learning rate of 0.05 decayed by 0.1 every 30 epochs after 150 epochs. The balancing parameters $\gamma_{kd}$ and $\alpha_{kd}$ are 0.1 and 0.9 for all settings, respectively. In Table \ref{test_accuracy_table_on_cifar100}, we report our results of four different settings as follows, distilled models without mixup (No Mixup), distilled models with PMU (10\% and 50\% when $\alpha=1$), and distilled models with full mixup (FMU, $\alpha=1$). As seen in Table \ref{test_accuracy_table_on_cifar100}, the students trained with PMU consistently outperform the ones trained without mixup. Surprisingly, in some cases (e.g., T:W40-2 \& S:W16-2 and T:W40-2 \& S:SN1), our student trained with PMU performs better than the teacher. 

\noindent \textbf{Experiments on ImageNet:}
Table \ref{test_accuracy_table_on_imagenet} shows the top-1 accuracy on ImageNet \cite{deng2009imagenet}. In this experiment, we choose ResNet34 and ResNet18 \cite{he2016deep} as the teacher network and student network, respectively. We train the model for 100 epochs and with initial learning rate is 0.1 decayed by 0.1 at 30, 60, and 80 epochs. The batch size is set as 256. For comparison with other distillation methods, the hyper-parameters of other methods follow their respective papers. The balancing parameters $\gamma_{kd}$ and $\alpha_{kd}$ are 0.1 and 0.9 respectively, and we report the partial mixup of 10\% and 100\% (FMU). We observe that the proposed method with full mixup boosts the top-1 and top-5 accuracy by 2.07\% and 1.56\% over the baseline and fully mixup shows better performance than 10\% partial mixup. We will discuss lower performance cases in the following section. 

\noindent\textbf{Performance analysis: } Based on Table \ref{test_accuracy_table_on_cifar100}, in some of the teacher-student combinations, we observe that adding more mixup pairs helps the student achieve higher accuracy than small amounts of pairs. Furthermore, for different architectural styles, the use of full mixup outperforms others in some cases. The presumption from this observation is that the networks from different architectures try to seek their solution paths, which means that the teacher and the student have unalike distributions in the logits, and thus imposing extra smoothness induced by strong augmentation might provide the student with additional information about how the different style teacher represents knowledge. Also, in the case of ImageNet having a large number of classes, the generated knowledge learned by two-mixing images and their labels might produce relatively less informative knowledge compared to CIFAR-100, so a strong smoothness by full mixup is favorable to distilling a better student model in this case. While we heuristically present that the performance is controlled by the amount of smoothness, how much exact smoothness should be imposed across datasets or networks remains an open question that can form the basis for further future work. 

\begin{table}[ht!]
\vskip -0.00in
\caption{Classification accuracy against the white-box attack with various perturbations of $\epsilon$ for every pixel. All methods are trained with full mixup ($\alpha=0.2$) and our distillation is trained with partial mixup only 10\% ($\alpha=1.0$) on CIFAR100.}\label{accuracy_on_adversarial_examples}
\vspace{-0.1in}
\begin{center}
\scalebox{0.67}{\begin{tabular}{|l|ccccc|}
\toprule
 & \multicolumn{5}{|c|}{FGSM} \\
\midrule
eps & Vanilla(+\textit{mixup}) & KD(+\textit{mixup}) & ICKD(+\textit{mixup}) & CRD(+\textit{mixup}) & Ours \\
\midrule
0.0 & 69.14(69.42) & 70.36(70.65) & 71.00(70.36) & 71.03(69.25) & \textbf{72.01} \\
0.001 & 61.14(63.18) & 64.25(65.02) & 64.74(64.59) & 64.09(63.98) & \textbf{68.92} \\
0.003 & 48.80(51.92) & 52.98(54.98) & 53.35(54.43) & 52.91(53.25) & \textbf{62.21} \\
0.005 & 39.00(42.59) & 44.13(46.28) & 44.00(46.40) & 43.99(44.20) & \textbf{55.89} \\
0.01 & 23.25(27.61) & 29.89(32.29) & 29.50(31.96) & 29.75(29.09) & \textbf{44.10} \\
\hline
& \multicolumn{5}{|c|}{I-FGSM}\\
\hline
0.0 & 69.14(69.42) & 70.36(70.65) & 71.00(70.36) & 71.03(69.25) & \textbf{72.01}\\
0.001 & 61.57(63.03) & 64.08(64.85) & 64.58(64.45) & 63.91(63.92) & \textbf{68.91}\\
0.003 & 46.79(50.55) & 51.43(53.56) & 51.72(53.11) & 51.39(51.80) & \textbf{61.53}\\
0.005 & 34.05(39.19) & 39.96(42.99) & 39.94(42.97) & 40.02(40.78) & \textbf{53.93}\\
0.01 & 13.28(18.58) & 19.71(23.12) & 19.34(23.13) & 19.59(19.70) & \textbf{37.40}\\
\hline
\end{tabular}}
\end{center}
\vspace{-3.0em} 
\end{table}

\begin{figure*}[ht!]
\centering
\includegraphics[width=0.92\linewidth]{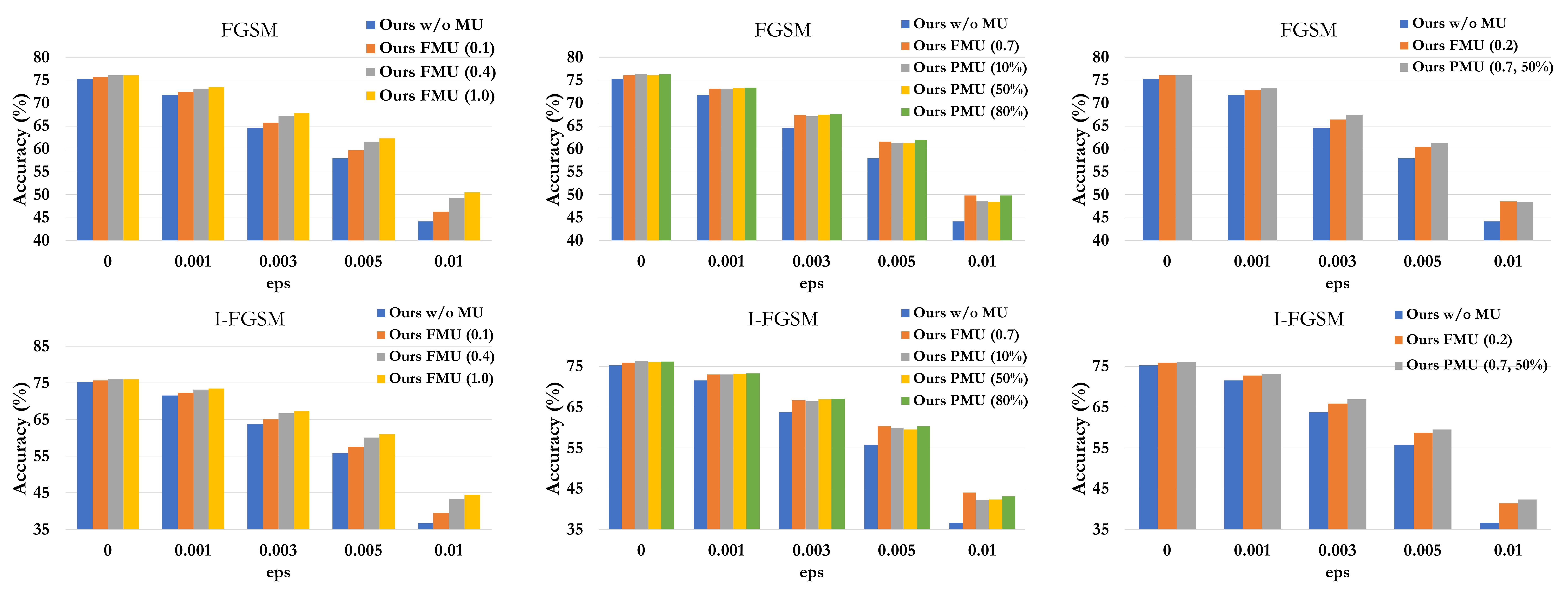}
\caption{Histogram of test accuracy under the attacks on CIFAR100. The first column indicates comparative results of different $\alpha$ when full mixup is used. The second column shows the comparison between full mixup and partial mixup with $\alpha=0.7$. And the last figure shows the comparison between FMU (0.2) and PMU (0.7, 50\%).}\label{Abl_std_Attack}
\vskip -0.15in
\end{figure*}

\subsection{Robustness to adversarial examples}\label{robustness_to_adversarial}
One undesirable consequence of models trained is their fragility to adversarial examples \cite{goodfellow2014explaining}. Adversarial examples are made by adding tiny (visually imperceptible) perturbations to legitimate samples to deteriorate the model performance. Unfortunately, many distillation methods have evolved to improve the performance of KD, disregarding attacks. Thus, in this section, we evaluate our model against white-box attacks where we used trained models themselves to generate adversarial examples using two common attacks, the Fast Gradient Sign Method (FGSM) and the Iterative FGSM (I-FGSM) methods \cite{goodfellow2014explaining}. For I-FGSM, we use 10 iterations with an equal step size. The results for both attacks are summarized in Table \ref{accuracy_on_adversarial_examples}. For distillation methods, the setting of teacher and student is T:RN110 \& S:RN20, where they were trained on CIFAR100. We applied mixup augmentation to all methods we explored (see. method+\textit{mixup} in the table). We trained our model using a partial mixup of only 10\% with $\alpha=1$. Even with only 10\% mixup pairs used, it shows impressively resistant to both attacks. We notice that distillation methods with the mixup, which utilize feature maps such as ICKD and CRD underperform those without mixup. We will show more comparative results of different amounts of mixup pairs in the next section.

\subsection{Ablation Study}\label{Ablation_study_section} In this section, we conduct an ablation study for hyper-parameters with a combination of T:WN40-2 \& S:WN16-2 networks. As shown in Table \ref{ablation_study_table}, the distilled models with PMU and high $\alpha$ value generally yields better performance. Further, to investigate how much the degree of partial mixup and $\alpha$ react against both attacks, we also show test accuracy in Figure \ref{Abl_std_Attack}. As shown in the first and second columns of the histogram, when strong augmentation is involved in training such as full mixup (FMU with a high value of $\alpha$) or the high number of mixup pairs (PMU 80\%), it improves robustness. Interestingly, only 10\% partial mixup (gray bar in the middle column) defends both attacks well. For the last column of the figure, we selected two distilled models from FMU ($\alpha=0.2$) and PMU (50\%, $\alpha=0.7$) where they have similar test performance in Table \ref{ablation_study_table}. We observe that PMU (50\%, $\alpha=0.7$) shows slightly higher robustness than FMU ($\alpha=0.2$).

\begin{table}[h!]
\vspace{-0.0in}
\begin{center}
\caption{Test accuracy under different settings of $\alpha$ and partial mixup amount on CIFAR100. The accuracy is averaged over $3$ runs. M. denotes mixup.}
\label{ablation_study_table}
\vspace{0.05in}
\scalebox{0.80}{\begin{tabular}{c|l|ccccc}
\toprule
\multirow{6}{*}{\rotatebox[origin=c]{90}{\scriptsize{w/ M.}}} & $\alpha$ & 0.1 & 0.2 & 0.4 & 0.7 & 1.0 \\ \cline{2-7} 
& PMU=10\% &  75.30 & 75.45 & 75.89 & 75.94 & 75.92  \\
& PMU=30\% &  75.51 & 75.40 & 75.71 & \textbf{76.21} & 75.95 \\
& PMU=50\% & 75.50 & 75.50 & 75.83 & 76.02 & \textbf{75.97} \\
& PMU=80\% & 75.36 & 75.66 & \textbf{76.05} & 76.19 & 75.78 \\
& FMU & \textbf{75.69} & \textbf{76.01} & 75.96 & 75.92 & 75.75 \\ \midrule
\multirow{2}{*}{\rotatebox[origin=c]{90}{\scriptsize{ w/o M.}}} & KD & & & 74.60\%\\
& Vanilla & & & 73.26\%\\
\bottomrule
\end{tabular}}
\end{center}
\vskip -0.40in
\end{table}

\section{Conclusions}
In this work, we study the role of a mixup in knowledge distillation. We observed that a mixup-trained teacher network produces inferior supervision due to exorbitant smoothness imposed on the features and logits, especially at high temperatures during distillation. Thus, the students experience a reduction in their performance in KD. We support our findings through a series of empirical analyses and large-scale experiments on the image classification task. Our findings provide insight into the inner workings of the distilled model trained with mixup augmentation. These insights allow us to develop an improved learning strategy using rescaled logits and partial mixups.

As we mentioned earlier, various mix-based augmentations have shown their effectiveness for particular tasks. However, these augmentations may tend to create unreasonable training samples as it blends random images \cite{kim2021co}, which could distort reasonable relative structure among categories. As a result, this potentially yields an unfavorable-smooth effect on logits during distillation. Therefore, developing an augmentation method that automatically selects more reasonable samples ensuring the best-fitting smoothness would further boost the progress in the distillation field. We will further develop this technique in our future work.

\section{Acknowledgements}
This material is based upon work supported by the Defense Advanced Research Projects Agency (DARPA) under Agreement No. HR00112290073. Approved for public release; distribution is unlimited.

{\small
\bibliographystyle{ieee_fullname}
\bibliography{egbib}
}

\end{document}